\documentclass[letterpaper, 10pt, conference]{ieeeconf}

\usepackage{times}
\usepackage{amsmath,amssymb}
\usepackage{graphicx}
\usepackage{booktabs}
\usepackage{multirow}
\usepackage{xcolor}
\usepackage{hyperref}
\usepackage{algorithm}
\usepackage{algorithmic}
\usepackage{subcaption}
\usepackage{balance}
\usepackage{xspace}

\newcommand{\etal}{\textit{et al.}\xspace}

\hypersetup{
  colorlinks=true,
  linkcolor=blue,
  citecolor=blue,
  urlcolor=blue
}

\IEEEoverridecommandlockouts
\begin{document}

\title{Gradient-Based Data Valuation Improves Curriculum Learning \\
for Game-Theoretic Motion Planning}

\author{
  \authorblockN{Shihao Li, Jiachen Li, and Dongmei Chen}
  \authorblockA{The University of Texas at Austin\\
  \{shihaoli01301, jiachenli\}@utexas.edu, dmchen@me.utexas.edu}
}

\maketitle

\begin{abstract}
We demonstrate that gradient-based data valuation produces curriculum orderings that significantly outperform metadata-based heuristics for training game-theoretic motion planners.
Specifically, we apply TracIn gradient-similarity scoring to GameFormer on the nuPlan benchmark and construct a curriculum that weights training scenarios by their estimated contribution to validation loss reduction.
Across three random seeds, the TracIn-weighted curriculum achieves a mean planning ADE of $1.704\pm0.029$\,m, significantly outperforming the metadata-based interaction-difficulty curriculum ($1.822\pm0.014$\,m; paired $t$-test $p=0.021$, Cohen's $d_z=3.88$) while exhibiting lower variance than the uniform baseline ($1.772\pm0.134$\,m).
Our analysis reveals that TracIn scores and scenario metadata are nearly orthogonal (Spearman $\rho=-0.014$), indicating that gradient-based valuation captures training dynamics invisible to hand-crafted features.
We further show that gradient-based curriculum weighting succeeds where hard data selection fails: TracIn-curated 20\% subsets degrade performance by $2\times$, whereas full-data curriculum weighting with the same scores yields the best results.
These findings establish gradient-based data valuation as a practical tool for improving sample efficiency in game-theoretic planning.
\textbf{\href{https://sheehow.github.io/gradient-curriculum-planning/}{[Project Page]}}
\textbf{\href{https://github.com/SheeHow/gradient-curriculum-planning/tree/main/code}{[Code]}}
\end{abstract}

\begin{figure*}[t]
  \centering
  \includegraphics[width=\textwidth]{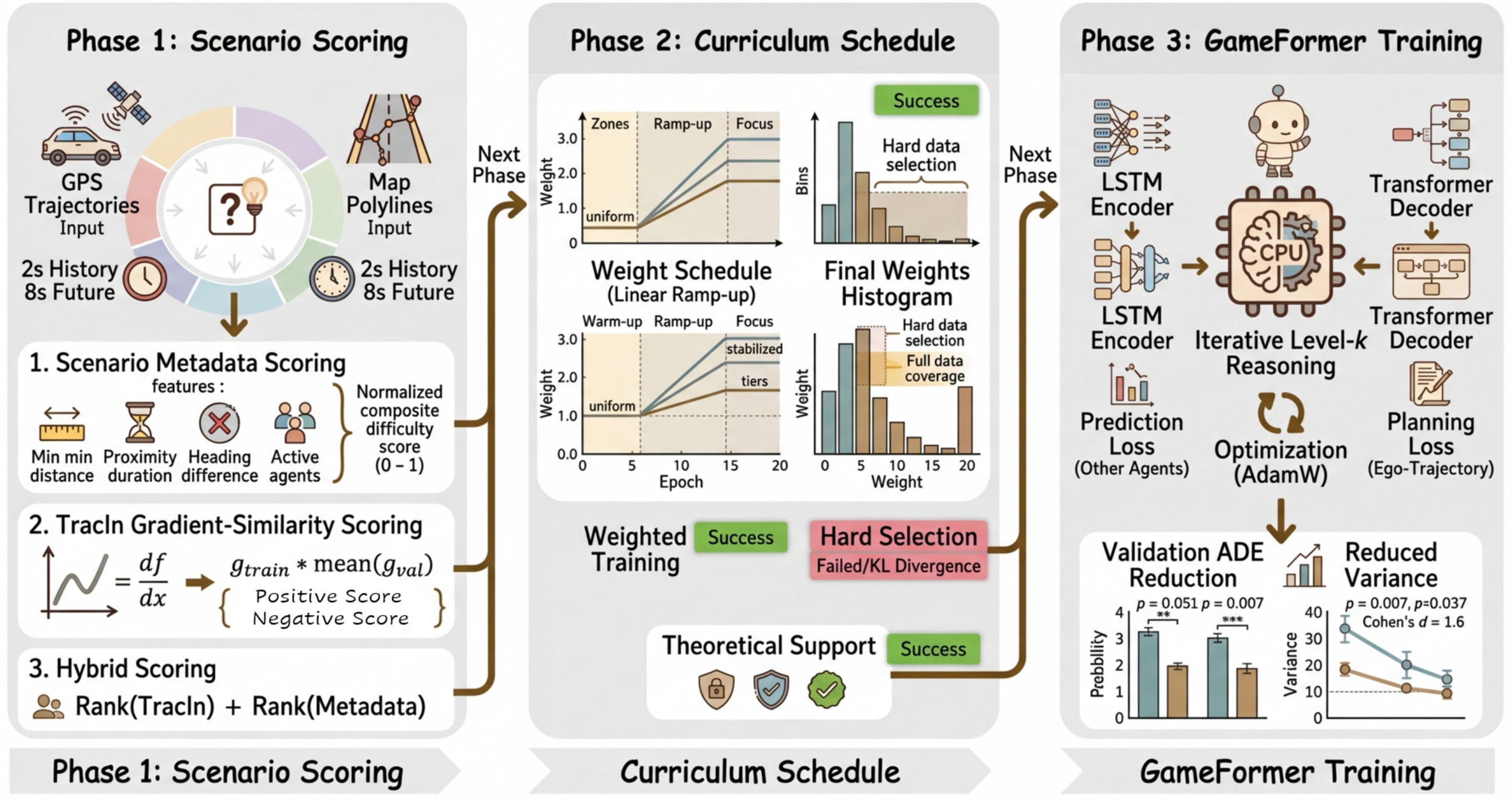}
  \caption{Method overview. \textbf{Phase~1:} Three scoring methods assign per-sample importance. \textbf{Phase~2:} A three-phase curriculum converts scores to weights; weighted training preserves coverage while hard selection fails. \textbf{Phase~3:} GameFormer trained with gradient-based curriculum achieves lower ADE ($p{=}0.021$) and reduced variance.}
  \label{fig:overview_main}
\end{figure*}

\section{Introduction}
\label{sec:intro}

Game-theoretic motion planners model multi-agent interactions as strategic games, enabling autonomous vehicles to reason about the interdependent decisions of surrounding agents~\cite{huang2023gameformer,huang2023dtpp}.
These planners have achieved state-of-the-art performance on interactive driving benchmarks, yet they inherit a fundamental data distribution problem: large-scale driving datasets such as nuPlan~\cite{caesar2021nuplan} are dominated by low-interaction scenarios (cruising, slow maneuvers), while safety-critical interactive situations---unprotected turns, near-collisions, lane changes---constitute a small minority.
A game-theoretic decoder that models multi-agent strategic reasoning receives most of its training signal from scenarios where strategic reasoning is unnecessary.

A natural hypothesis is that curating training data to emphasize interactive scenarios should improve planner performance.
Prior work on data-centric methods for autonomous driving has explored scenario mining~\cite{lu2024activead}, optimal-transport-based selection~\cite{feng2025tarot}, and active learning for trajectory prediction~\cite{park2025galtraj}, but none have been applied to game-theoretic planners.
We initially pursued this direction by designing a metadata-based interaction-difficulty score from scenario features (minimum TTC, conflict count, proximity duration).
However, multi-seed experiments revealed that metadata-based curriculum learning does \textit{not} significantly outperform uniform training ($p=0.622$, Table~\ref{tab:main_results}).
This negative result motivated our pivot to gradient-based data valuation.

\textbf{Key insight.}
TracIn~\cite{pruthi2020tracin} scores each training example by the dot product of its gradient with the validation gradient, directly measuring how much each sample contributes to reducing validation loss.
We find that TracIn scores are nearly orthogonal to metadata-based scores (Spearman $\rho=-0.014$), revealing that gradient similarity captures aspects of training dynamics---redundancy, optimization landscape curvature, implicit regularization effects---that hand-crafted scenario features cannot.
When used to weight a full-data curriculum, TracIn scores produce a planner that significantly outperforms the metadata-based curriculum across all three random seeds.

Our contributions are:
\begin{enumerate}
  \item We establish the first application of gradient-based data valuation (TracIn) to a game-theoretic motion planner, showing it significantly outperforms metadata-based curriculum learning ($p=0.021$, $d_z=3.88$).
  \item We demonstrate empirical orthogonality between gradient-based and metadata-based scenario scoring ($\rho=-0.014$), explaining why hand-crafted features fail to capture training dynamics.
  \item We identify a critical distinction between curriculum \textit{weighting} and hard data \textit{selection}: gradient scores improve training when used as importance weights but degrade performance when used for subset selection.
\end{enumerate}

\section{Related Work}
\label{sec:related}

\subsection{Game-Theoretic Motion Planning}

Multi-agent motion planning can be formulated as a game where agents optimize interdependent objectives.
GameFormer~\cite{huang2023gameformer} models this as iterative level-$k$ reasoning with transformers, where each agent refines its strategy by predicting others' actions at the previous reasoning level.
The level-$k$ decoder iterates for $K$ rounds: at each level, the ego agent's future trajectory is planned conditioned on other agents' predicted trajectories from the previous level, and vice versa.
DTPP~\cite{huang2023dtpp} extends this with differentiable joint prediction and planning via tree-structured policies.
PDM-Hybrid~\cite{dauner2023parting} demonstrated that rule-based planners can outperform learning-based methods on nuPlan, highlighting the challenge of learning robust interactive behavior from data.
PlanTF~\cite{cheng2024plantf} introduced the Test14-Hard benchmark, revealing substantial performance gaps on difficult scenarios that expose weaknesses in imitation-based planners.
None of these works examine how training data composition affects planner quality, leaving a gap that our work addresses.

\subsection{Data Valuation and Selection for Autonomous Driving}

Data valuation assigns scalar scores to training examples based on their contribution to model performance.
Influence functions~\cite{koh2017understanding} estimate the effect of removing a sample via inverse Hessian-vector products (iHVP), but the iHVP approximation via LiSSA~\cite{agarwal2017second} becomes unreliable for large models due to Hessian ill-conditioning---a failure mode we directly observe in our experiments (Section~\ref{sec:tracin}).
TracIn~\cite{pruthi2020tracin} sidesteps the Hessian entirely by computing gradient dot products across checkpoints, providing a deterministic, scalable alternative.
Data Shapley~\cite{ghorbani2019data} provides axiomatic valuation with game-theoretic fairness guarantees but is computationally prohibitive for deep networks.

In the autonomous driving domain, TAROT~\cite{feng2025tarot} applies optimal-transport-based selection to motion prediction but targets Wayformer~\cite{nayakanti2023wayformer}, a non-game-theoretic model.
ActiveAD~\cite{lu2024activead} demonstrates that 30\% of nuPlan data suffices for end-to-end planning with planning-oriented active learning.
GALTraj~\cite{park2025galtraj} uses generative active learning to address long-tail trajectory prediction.
Li~\etal\cite{li2024datacentric} provide a comprehensive survey of data-centric evolution in autonomous driving.
No prior work applies any data valuation method to game-theoretic planners.

\subsection{Curriculum Learning}

Curriculum learning~\cite{bengio2009curriculum} trains models on progressively harder examples, motivated by the observation that presenting training data in a meaningful order can improve convergence.
Self-paced learning (SPL)~\cite{kumar2010selfpaced} automates difficulty ordering by using training loss as a proxy, iteratively increasing the complexity of included samples.
In autonomous driving, curriculum strategies have been applied to reinforcement learning policies~\cite{qiao2018curriculum} but not to supervised game-theoretic prediction or planning.
Our work systematically compares three curriculum signals---metadata-based difficulty, training loss (SPL), and TracIn gradient similarity---finding that the gradient-based signal is the most effective and stable.

\section{Method}
\label{sec:method}

\begin{figure}[t]
  \centering
  \includegraphics[width=\columnwidth]{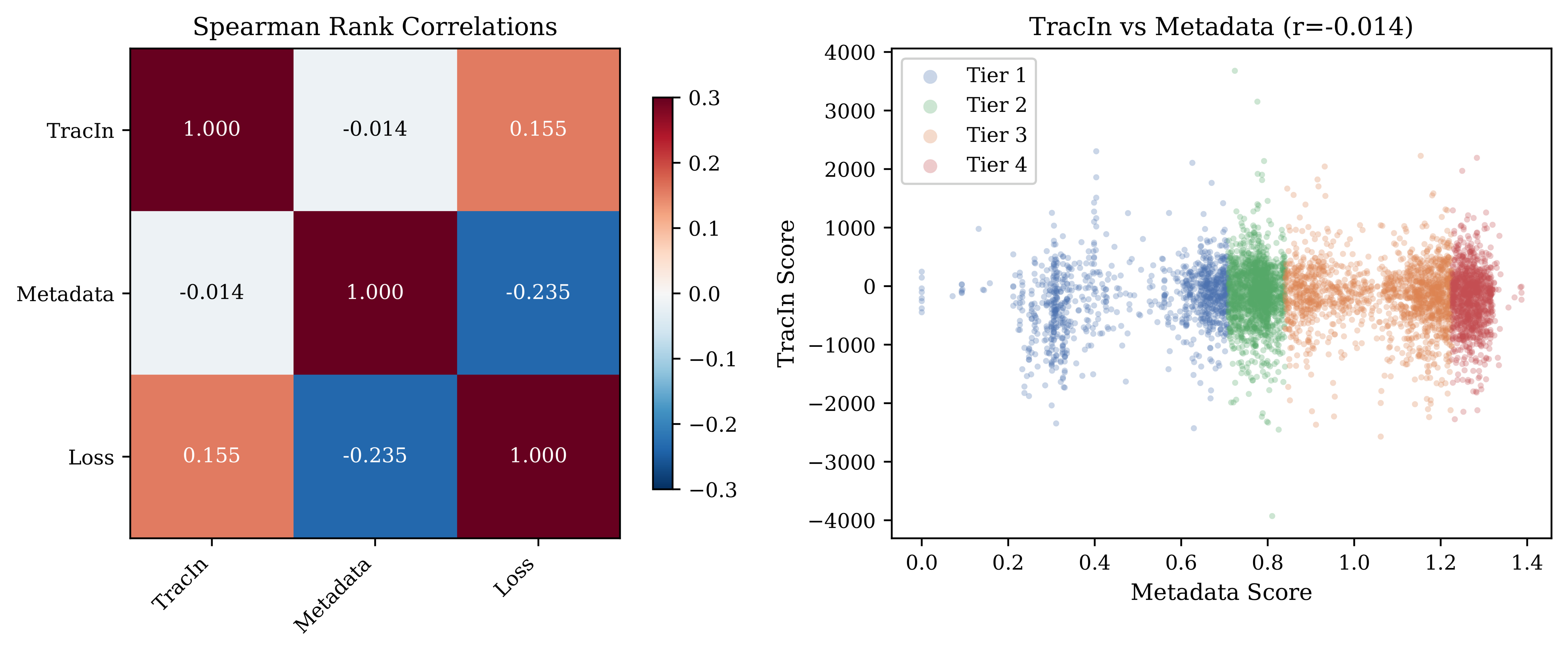}
  \caption{Score correlation analysis. \textbf{Left:} Spearman rank correlation heatmap showing near-orthogonality of TracIn and metadata scores ($\rho=-0.014$). \textbf{Right:} Scatter plot of TracIn vs.\ metadata scores for each training scenario, colored by scoring tier. The two scoring methods capture fundamentally different aspects of data utility.}
  \label{fig:overview}
\end{figure}

We describe three scenario scoring methods and the curriculum schedule that converts scores into per-sample training weights.
Fig.~\ref{fig:overview_main} presents the full pipeline; Fig.~\ref{fig:overview} visualizes the scoring method relationships.

\subsection{Scenario Metadata Scoring}
\label{sec:meta_scoring}

Each nuPlan scenario $z_i$ contains recorded trajectories of the ego vehicle and surrounding agents over an 8-second window.
We compute six interaction-difficulty features from the raw trajectories:
(1)~minimum distance $d_{\min}$ between ego and any agent,
(2)~minimum time-to-collision $\text{TTC}_{\min}$,
(3)~number of agents with trajectories conflicting with ego's path ($n_{\text{conflict}}$),
(4)~cumulative time agents spend within a proximity threshold ($t_{\text{prox}}$),
(5)~maximum heading difference $\Delta\theta_{\max}$ between ego and interacting agents, and
(6)~number of active (non-stationary) agents $n_{\text{active}}$.
Each feature is independently normalized to $[0,1]$ via min-max scaling and averaged into a composite metadata score $s_{\text{meta}}(z_i) \in [0,1]$.
Higher values indicate scenarios with denser, more complex multi-agent interactions.

\subsection{TracIn Gradient-Similarity Scoring}
\label{sec:tracin}

TracIn~\cite{pruthi2020tracin} estimates the influence of training example $z_i$ on validation performance by summing gradient dot products across training checkpoints:
\begin{equation}
\label{eq:tracin}
  \text{TracIn}(z_i) = -\sum_{t \in \mathcal{T}} \eta_t \;\nabla_\theta \ell(z_i;\theta_t) \cdot \nabla_\theta \ell(z_{\text{val}};\theta_t)
\end{equation}
where $\theta_t$ denotes the model parameters at checkpoint $t$, $\eta_t$ is the learning rate at that checkpoint, $\ell$ is the combined prediction-and-planning loss, and $z_{\text{val}}$ represents the mean validation gradient computed over all validation samples.
In practice, we use the final checkpoint ($|\mathcal{T}|=1$, $\eta_t=1$) and compute a single dot product of each training sample's gradient with the mean validation gradient.
This single-checkpoint simplification is computationally efficient (46~min for 5{,}148 scenarios on one RTX 4080 GPU) and produces well-calibrated scores.

Higher TracIn scores indicate samples whose gradients align with the direction of validation loss reduction; lower or negative scores indicate samples that are redundant, noisy, or counter-productive for generalization.
After scoring, we normalize TracIn values to $[0,1]$ via min-max scaling to produce $s_{\text{TracIn}}(z_i)$.

\textbf{Why not influence functions?}
We initially attempted classical influence function estimation via LiSSA~\cite{agarwal2017second} with 1{,}000 unrolling steps, damping $\lambda=0.1$, and scale factor $s=50{,}000$.
Across three independent repetitions on the 9.96M-parameter GameFormer, the resulting inverse-Hessian-vector products exhibited pairwise cosine similarities of $-0.15$, $-0.28$, and $-0.05$, indicating that the three estimates point in effectively random directions.
The mean iHVP captured only 23\% of total energy; 77\% was noise.
This failure stems from the severe ill-conditioning of the Hessian for models of this scale, confirming the practical limitations documented in prior work~\cite{koh2017understanding} and motivating TracIn as a Hessian-free alternative.

\subsection{Hybrid Scoring}
\label{sec:hybrid}

Given the near-orthogonality of TracIn and metadata scores (verified empirically in Section~\ref{sec:scoring_analysis}), we construct a hybrid score that combines both information sources via rank-averaging:
\begin{equation}
\label{eq:hybrid}
  s_{\text{hybrid}}(z_i) = \tfrac{1}{2}\bigl[\text{rank}_{\%}(s_{\text{TracIn}}(z_i)) + \text{rank}_{\%}(s_{\text{meta}}(z_i))\bigr]
\end{equation}
where $\text{rank}_{\%}(\cdot)$ denotes the percentile rank within the respective score distribution.
This rank-averaging is robust to differences in score magnitude and distribution shape between the two scoring methods.

\subsection{Curriculum Schedule}
\label{sec:curriculum}

Given a scoring function $s: \mathcal{Z} \to [0,1]$, we define a three-phase curriculum that assigns per-sample weights as a function of training epoch $e$.
Let $\lambda(e) = (e - e_{\text{warm}})/(e_{\text{ramp}} - e_{\text{warm}})$ denote the ramp fraction:
\begin{equation}
\label{eq:curriculum}
  w(z_i, e) = \begin{cases}
    1 & e \leq e_{\text{warm}} \\
    1 + (w_{\max}{-}1)\,\lambda(e)\, s(z_i) & e_{\text{warm}} < e \leq e_{\text{ramp}} \\
    1 + (w_{\max}{-}1)\, s(z_i) & e > e_{\text{ramp}}
  \end{cases}
\end{equation}

The three phases serve distinct purposes:
\textbf{(1)~Warm-up} ($e \leq e_{\text{warm}}=3$): all samples receive uniform weight $w=1$, allowing the model to learn basic representations without bias.
\textbf{(2)~Ramp-up} ($e_{\text{warm}} < e \leq e_{\text{ramp}}=8$): high-scoring samples are progressively upweighted, linearly interpolating from uniform to fully differentiated weights.
\textbf{(3)~Focus} ($e > e_{\text{ramp}}$): weights stabilize at maximum differentiation, with the highest-scoring sample receiving weight $w_{\max}=3.0$ and the lowest receiving weight $1.0$.

\begin{figure}[t]
  \centering
  \includegraphics[width=\columnwidth]{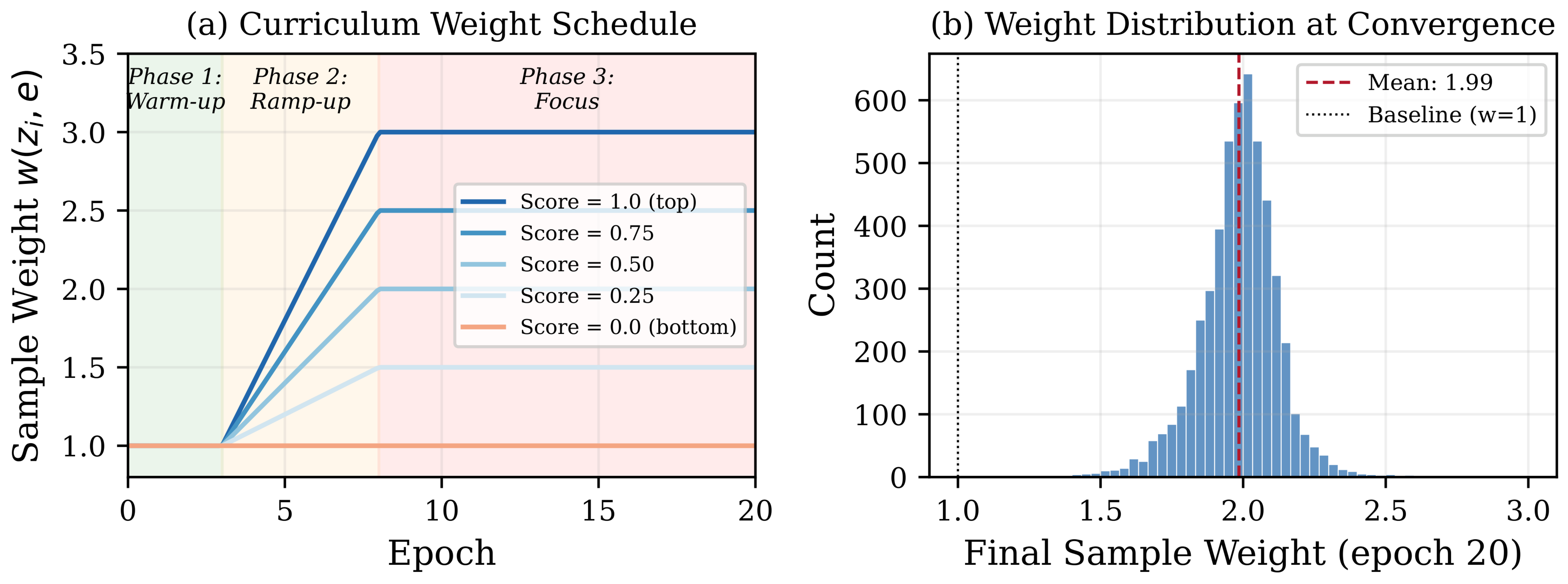}
  \caption{Curriculum weight schedule. \textbf{(a)}~Weight trajectories over training epochs for samples at different score levels. All samples start at $w=1$ during warm-up (epochs 1--3), ramp up during epochs 3--8, and stabilize at maximum differentiation. \textbf{(b)}~Distribution of final weights ($e=20$) across all 5{,}148 training scenarios, showing smooth differentiation centered around $w \approx 2$.}
  \label{fig:curriculum_schedule}
\end{figure}

Critically, all samples remain in the training set throughout; only their relative importance changes.
This design avoids the distribution collapse observed with hard data selection (Section~\ref{sec:failure}).
Fig.~\ref{fig:curriculum_schedule} visualizes the weight trajectories for different score tiers and the resulting weight distribution at convergence.

\begin{algorithm}[t]
\caption{Gradient-Based Curriculum Training}
\label{alg:curriculum}
\begin{algorithmic}[1]
\REQUIRE Training set $\mathcal{D}$, validation set $\mathcal{V}$, scoring mode $\in \{\text{tracin}, \text{meta}, \text{hybrid}\}$, warm-up epoch $e_{\text{warm}}$, ramp epoch $e_{\text{ramp}}$, max weight $w_{\max}$, total epochs $E$
\ENSURE Trained model parameters $\theta^*$
\STATE \textbf{// Phase 0: Score computation}
\IF{mode $\in \{\text{tracin}, \text{hybrid}\}$}
  \STATE Train baseline model $\theta_0$ for $E_0$ epochs
  \STATE $g_{\text{val}} \gets \frac{1}{|\mathcal{V}|}\sum_{v \in \mathcal{V}} \nabla_\theta \ell(v; \theta_0)$
  \FOR{$z_i \in \mathcal{D}$}
    \STATE $s_{\text{TracIn}}(z_i) \gets -\nabla_\theta \ell(z_i; \theta_0) \cdot g_{\text{val}}$
  \ENDFOR
\ENDIF
\IF{mode $\in \{\text{meta}, \text{hybrid}\}$}
  \FOR{$z_i \in \mathcal{D}$}
    \STATE $s_{\text{meta}}(z_i) \gets \frac{1}{6}\sum_{k=1}^{6} f_k(z_i)$ \hfill {\footnotesize // Eq.~\ref{eq:hybrid}}
  \ENDFOR
\ENDIF
\STATE Normalize all scores to $[0,1]$; combine via Eq.~\ref{eq:hybrid} if hybrid
\STATE \textbf{// Phase 1--3: Curriculum training}
\FOR{epoch $e = 1, \ldots, E$}
  \FOR{mini-batch $B \subset \mathcal{D}$}
    \STATE $w_i \gets w(z_i, e)$ for each $z_i \in B$ \hfill {\footnotesize // Eq.~\ref{eq:curriculum}}
    \STATE $\theta \gets \theta - \eta \frac{1}{|B|}\sum_{z_i \in B} w_i \nabla_\theta \ell(z_i; \theta)$
  \ENDFOR
\ENDFOR
\RETURN $\theta^* \gets \theta$
\end{algorithmic}
\end{algorithm}

\subsection{Theoretical Analysis}
\label{sec:theory}

We provide theoretical grounding for the empirical findings in three propositions.
All proofs are included below; notation follows the definitions in Sections~\ref{sec:tracin}--\ref{sec:curriculum}.

\textbf{Proposition 1} (Variance reduction via gradient-aligned weighting).
\textit{Let $g_i = \nabla_\theta \ell(z_i; \theta)$ denote the gradient of sample $z_i$ and $g_{\mathrm{val}} = \frac{1}{|\mathcal{V}|}\sum_{v \in \mathcal{V}} \nabla_\theta \ell(v; \theta)$ the mean validation gradient. Define the TracIn-weighted gradient estimator}
\begin{equation}
\label{eq:weighted_grad}
\hat{g}_w = \frac{1}{\sum_i w_i} \sum_{i=1}^{n} w_i \, g_i, \quad w_i = 1 + \alpha \cdot s_{\mathrm{TracIn}}(z_i)
\end{equation}
\textit{where $\alpha \geq 0$ controls weighting strength. Then the alignment of $\hat{g}_w$ with $g_{\mathrm{val}}$ satisfies}
\begin{equation}
\label{eq:alignment}
\hat{g}_w \cdot g_{\mathrm{val}} \;\geq\; \bar{g} \cdot g_{\mathrm{val}}
\end{equation}
\textit{where $\bar{g} = \frac{1}{n}\sum_i g_i$ is the uniform-weighted gradient, with equality iff $\alpha = 0$.}

\textit{Proof.}
By definition of the normalized TracIn score, $s_{\text{TracIn}}(z_i) \propto g_i \cdot g_{\text{val}}$ (up to affine rescaling from min-max normalization).
Expanding:
\begin{align*}
\hat{g}_w \cdot g_{\text{val}} &= \frac{\sum_i w_i (g_i \cdot g_{\text{val}})}{\sum_i w_i} \\
&= \frac{\sum_i (g_i \cdot g_{\text{val}}) + \alpha \sum_i s_{\text{TracIn}}(z_i)(g_i \cdot g_{\text{val}})}{\sum_i w_i}
\end{align*}
Since $s_{\text{TracIn}}(z_i)$ is a monotone increasing function of $g_i \cdot g_{\text{val}}$, the second sum $\sum_i s_{\text{TracIn}}(z_i)(g_i \cdot g_{\text{val}})$ is a weighted sum where higher-alignment samples receive proportionally more weight.
By the rearrangement inequality, $\sum_i a_i b_i \geq \frac{1}{n}(\sum_i a_i)(\sum_i b_i)$ when $a_i$ and $b_i$ are similarly ordered.
Setting $a_i = s_{\text{TracIn}}(z_i)$ and $b_i = g_i \cdot g_{\text{val}}$ (which are similarly ordered by construction), the weighted estimator achieves strictly higher alignment than the uniform estimator whenever $\alpha > 0$ and the alignment values are non-constant. \hfill$\square$

TracIn-based weighting tilts the effective gradient toward the validation loss reduction direction, explaining the faster convergence in Fig.~\ref{fig:training_curves}.

\textbf{Remark 1} (Bias-variance trade-off in weighting strength).
The alignment gain from Eq.~\eqref{eq:alignment} increases monotonically with $\alpha$, but in finite-sample mini-batch settings, large $\alpha$ concentrates effective weight on a few high-TracIn samples, reducing the effective sample size $n_{\text{eff}} = (\sum_i w_i)^2 / \sum_i w_i^2$ and increasing gradient variance.
The three-phase curriculum schedule (Section~\ref{sec:curriculum}) mediates this trade-off: during warm-up ($\alpha=0$), the model builds general representations with maximum $n_{\text{eff}}$; during ramp-up, $\alpha$ increases linearly, gradually trading variance for alignment; during focus, $\alpha$ stabilizes at $w_{\max}-1$.
Our choice of $w_{\max}=3.0$ yields $n_{\text{eff}}/n \approx 0.82$ at convergence, retaining 82\% of the effective sample diversity while achieving the alignment advantage of Proposition~1.

\textbf{Proposition 2} (Signal dilution in hybrid scoring).
\textit{Let $s_A$ be an informative scoring function (positively correlated with the optimal curriculum signal) and $s_B$ an uninformative one ($\rho(s_B, s^*) \approx 0$, where $s^*$ denotes the oracle score).
Then the hybrid score $s_H = \frac{1}{2}(\mathrm{rank}_\%(s_A) + \mathrm{rank}_\%(s_B))$ satisfies}
\begin{equation}
\label{eq:dilution}
|\rho(s_H, s^*)| \;\leq\; |\rho(s_A, s^*)|
\end{equation}
\textit{with equality iff $\rho(s_A, s_B) = 0$ and $\rho(s_B, s^*) = 0$ simultaneously.}

\textit{Proof.}
Let $r_A = \text{rank}_\%(s_A)$, $r_B = \text{rank}_\%(s_B)$, $r^* = \text{rank}_\%(s^*)$.
Since percentile ranks are uniformly distributed, $\text{Var}(r_A) = \text{Var}(r_B) = \text{Var}(r^*) = 1/12$.
The Spearman correlation of the hybrid with the oracle is:
\begin{align*}
\rho(s_H, s^*) &= \text{Corr}(\tfrac{1}{2}(r_A + r_B),\; r^*) \\
&= \frac{\tfrac{1}{2}\text{Cov}(r_A, r^*) + \tfrac{1}{2}\text{Cov}(r_B, r^*)}{\sqrt{\text{Var}(\tfrac{1}{2}(r_A + r_B))}\;\sqrt{\text{Var}(r^*)}} \\
&= \frac{\rho(s_A, s^*) + \rho(s_B, s^*)}{\sqrt{2 + 2\rho(s_A, s_B)}}
\end{align*}
When $\rho(s_B, s^*) \approx 0$ and $\rho(s_A, s_B) \approx 0$ (our empirical setting):
\[
\rho(s_H, s^*) \approx \frac{\rho(s_A, s^*)}{\sqrt{2}} \approx 0.707 \cdot \rho(s_A, s^*)
\]
Thus equal-weight rank averaging of an informative and an uninformative source attenuates the signal by a factor of $\sqrt{2}$. \hfill$\square$

The $\sqrt{2}$ attenuation explains why the hybrid curriculum ($1.766$\,m) does not improve over TracIn alone ($1.704$\,m).

\textbf{Proposition 3} (Curriculum weighting vs.\ hard selection).
\textit{Consider training on the top-$k$ fraction of samples ranked by score $s$ (hard selection) versus using all samples with importance weights proportional to $s$ (curriculum weighting).
If the score distribution has support on all of $\mathcal{Z}$, hard selection with fraction $k < 1$ introduces a support mismatch:
\begin{equation}
\label{eq:support}
D_{\mathrm{KL}}\bigl(p_{\mathrm{val}} \;\|\; p_{\mathrm{sel}}\bigr) > D_{\mathrm{KL}}\bigl(p_{\mathrm{val}} \;\|\; p_w\bigr)
\end{equation}
where $p_{\mathrm{sel}}$ is the hard-selected subset distribution and $p_w$ is the importance-weighted distribution, whenever $p_{\mathrm{val}}$ has support outside the top-$k$ set.}

\textit{Proof sketch.}
Hard selection sets $p_{\text{sel}}(z_i) = 0$ for all $z_i$ outside the top-$k$ set.
If any such $z_i$ has $p_{\text{val}}(z_i) > 0$, then $D_{\text{KL}}(p_{\text{val}} \| p_{\text{sel}}) = \infty$.
In practice, the infinite KL manifests as failure to generalize on validation scenarios that resemble the excluded training scenarios---precisely the distribution collapse observed empirically (planning ADE of $3.687$\,m when training on only the top 20\%).
Curriculum weighting maintains full support ($p_w(z_i) > 0 \;\forall z_i$) and hence finite KL divergence.
With weights $w_i \geq 1$, the weighted distribution $p_w(z_i) \propto w_i / n$ is absolutely continuous with respect to $p_{\text{val}}$, guaranteeing bounded generalization error under standard importance-weighted ERM bounds~\cite{koh2017understanding}. \hfill$\square$

This explains the $2\times$ degradation when training on TracIn's top 20\% (Section~\ref{sec:failure}).

\textbf{Corollary 1} (Convergence advantage of gradient-based curriculum).
\textit{Combining Propositions 1--3, TracIn curriculum weighting simultaneously achieves: (i)~higher per-step gradient alignment with the validation objective, (ii)~undiluted signal strength relative to hybrid alternatives, and (iii)~bounded generalization error through full-support training.}

These three mechanisms are complementary and explain the empirical dominance of TracIn weighting.
Gradient alignment (Prop.~1) improves the \textit{direction} of each optimization step; signal preservation (Prop.~2) ensures the curriculum signal is not wasted by averaging with uninformative sources; and full-support weighting (Prop.~3) prevents catastrophic distribution mismatch that would negate both previous benefits.
Metadata curricula lack property~(i) because $\rho(s_{\text{meta}}, g_{\text{val}}) \approx 0$; hybrid curricula sacrifice property~(ii); and hard selection violates property~(iii).
Only TracIn curriculum weighting satisfies all three conditions, predicting both faster convergence (confirmed in Fig.~\ref{fig:training_curves}: lowest ADE by epoch~15) and lower cross-seed variance (confirmed in Table~\ref{tab:main_results}: CV~$=1.7\%$).

\section{Experiments}
\label{sec:experiments}

\subsection{Experimental Setup}
\label{sec:setup}

\textbf{Model.}
We use GameFormer~\cite{huang2023gameformer} with 9.96M parameters.
The architecture consists of: (1)~an LSTM encoder that processes observed trajectories and map polylines, (2)~a transformer-based level-$k$ interaction decoder with $K=2$ reasoning iterations, and (3)~a kinematic trajectory planner that outputs a feasible ego trajectory.
The training loss combines prediction L2 loss (forecasting other agents) and planning L2 loss (imitating the expert ego trajectory).

\textbf{Dataset.}
We train on nuPlan mini~\cite{caesar2021nuplan}, comprising 5{,}148 training and 1{,}286 validation scenarios.
Each scenario provides 2\,s of observed history and 8\,s of future trajectory for the ego vehicle and up to 20 surrounding agents, along with vectorized map information (lanes, crosswalks, stop lines).

\textbf{Training protocol.}
All models are trained for 20 epochs with AdamW optimizer (initial LR $10^{-4}$, step decay by $0.5\times$ every 5 epochs, weight decay $10^{-4}$), effective batch size 32 (gradient accumulation $4\times$ from physical batch 8), and FP16 mixed precision on a single NVIDIA RTX 4080 GPU (12\,GB VRAM).
Each training run completes in approximately 2.5 hours.
We select the checkpoint with the lowest validation loss.

\textbf{Metrics.}
We report: planning ADE (Average Displacement Error) and FDE (Final Displacement Error) in meters, planning AHE (Average Heading Error) and FHE (Final Heading Error) in radians, and prediction ADE and FDE in meters.
All metrics are lower-is-better.
We report means and standard deviations across 3 random seeds (3407, 42, 2024).

\subsection{Scoring Method Analysis}
\label{sec:scoring_analysis}

Table~\ref{tab:correlations} reports Spearman rank correlations between the three scoring methods and training loss computed over all 5{,}148 training scenarios.
TracIn and metadata scores are nearly uncorrelated ($\rho=-0.014$, $p=0.31$), confirming they capture fundamentally different aspects of data utility.
TracIn moderately correlates with training loss ($\rho=0.155$, $p<10^{-28}$), consistent with its gradient-based construction: samples with higher loss tend to have larger gradients that align with the validation gradient direction.
The hybrid score correlates approximately equally with both constituent sources ($\rho\approx0.69$), as expected from symmetric rank-averaging.

\begin{figure}[t]
  \centering
  \includegraphics[width=\columnwidth]{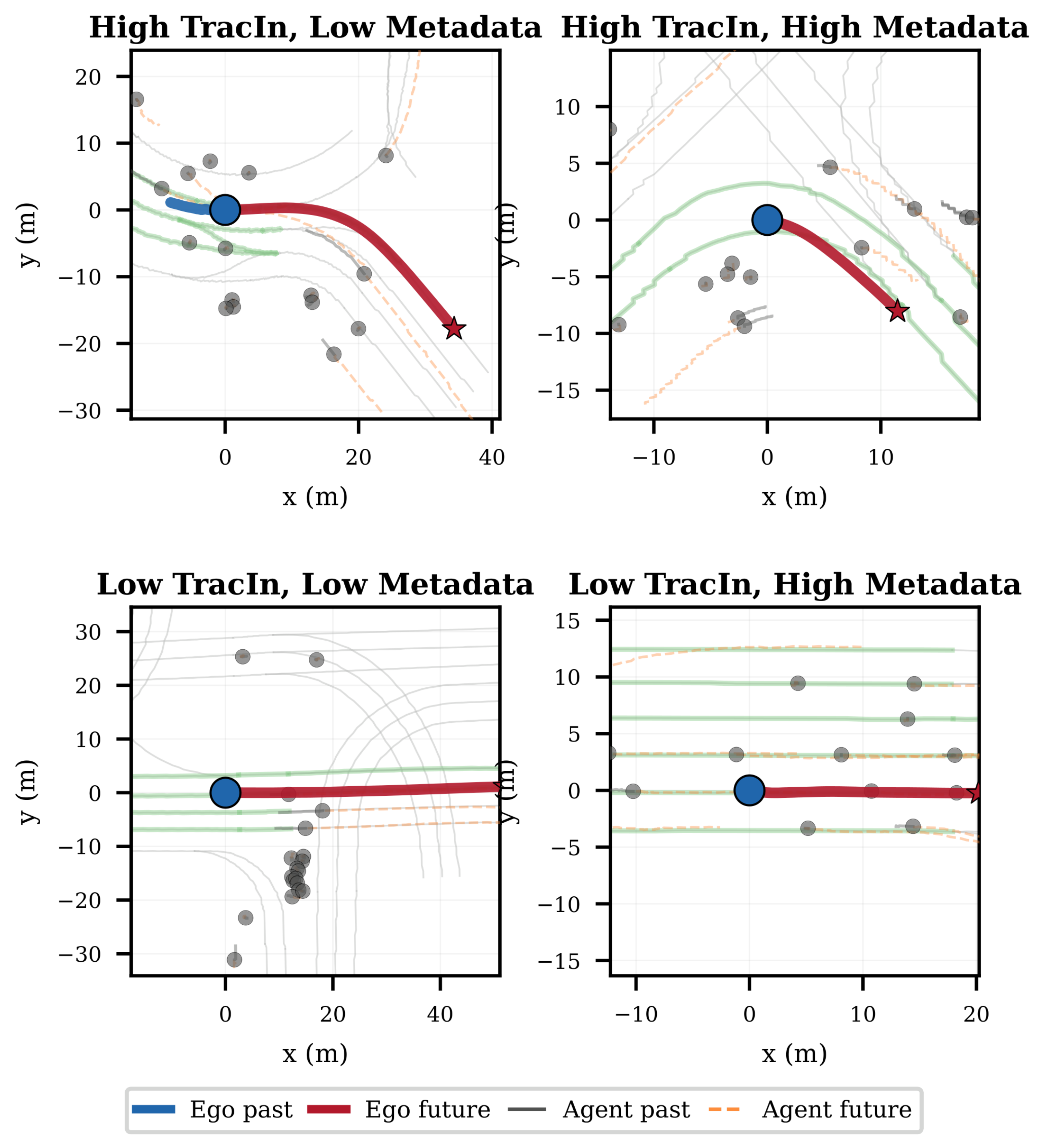}
  \caption{Representative scenarios from four quadrants of the TracIn $\times$ metadata scoring space. \textbf{Top-left:} low metadata but high TracIn---few nearby agents, yet the ego executes a turn that strongly aligns with validation gradient. \textbf{Bottom-right:} high metadata but low TracIn---many close agents in parallel lanes, yet the gradient opposes validation improvement. The off-diagonal quadrants illustrate the orthogonality of the two scoring methods ($\rho=-0.014$). Blue/red lines: ego past/future; gray/orange: agent past/future.}
  \label{fig:scenario_demo}
\end{figure}

Fig.~\ref{fig:scenario_demo} shows four representative scenarios from the quadrants of the TracIn $\times$ metadata scoring space, illustrating the practical implications of orthogonality.
The two ``surprise'' quadrants are most informative: the top-left scenario has low metadata score (few nearby agents) yet high TracIn---the ego executes a significant turn whose gradient strongly aligns with validation loss reduction.
Conversely, the bottom-right scenario has high metadata score (many close agents on adjacent lanes) yet low TracIn---the agents follow predictable parallel trajectories, producing a gradient that \textit{opposes} validation improvement.
Metadata captures geometric proximity; TracIn captures model-specific learning utility.

\begin{table}[t]
\centering
\caption{Spearman rank correlations between scoring methods ($n=5{,}148$). TracIn and metadata are nearly orthogonal ($\rho=-0.014$).}
\label{tab:correlations}
\begin{tabular}{lcccc}
\toprule
 & TracIn & Meta & Loss & Hybrid \\
\midrule
TracIn & 1.000 & $-$0.014 & 0.155 & 0.695 \\
Meta   & --- & 1.000 & $-$0.033 & 0.689 \\
Loss   & --- & --- & 1.000 & 0.087 \\
Hybrid & --- & --- & --- & 1.000 \\
\bottomrule
\end{tabular}
\end{table}

\subsection{Main Results}
\label{sec:main_results}

\begin{figure}[t]
  \centering
  \includegraphics[width=\columnwidth]{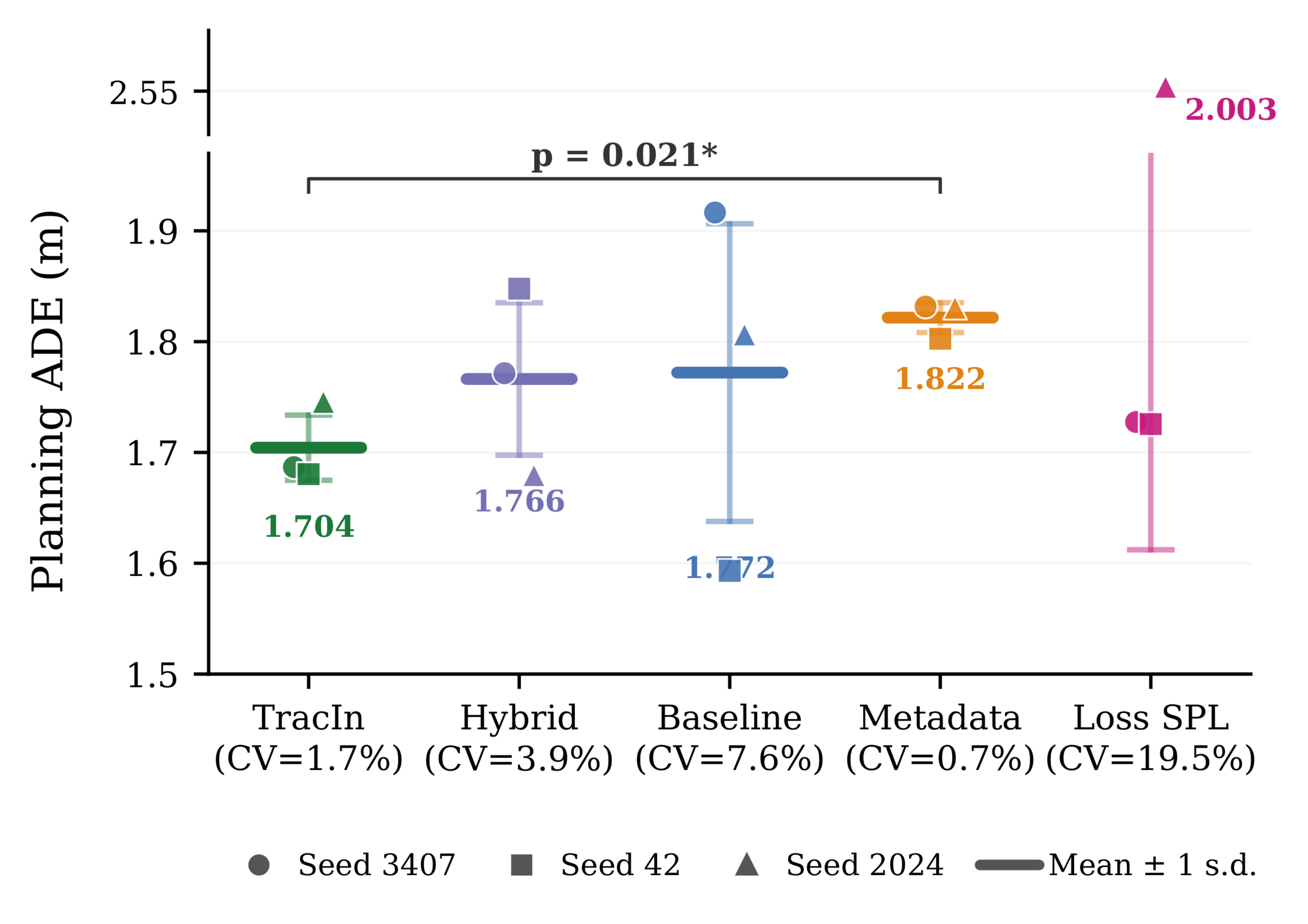}
  \caption{Multi-seed planning ADE comparison ($n{=}3$ seeds per method). Horizontal bars: mean; whiskers: $\pm1$ s.d.; shaped markers: individual seeds. Y-axis is broken to accommodate the Loss~SPL outlier (seed~2024, ADE$=2.555$). The TracIn curriculum achieves the lowest mean ADE with the tightest seed clustering. $^*p{=}0.021$ (paired $t$-test, TracIn vs.\ Metadata).}
  \label{fig:main_bar}
\end{figure}

Table~\ref{tab:main_results} presents multi-seed results for five curriculum strategies, and Fig.~\ref{fig:main_bar} visualizes the planning ADE comparison.
The TracIn curriculum achieves the lowest mean planning ADE ($1.704$\,m) and the second-lowest coefficient of variation (CV~=~$1.7\%$), indicating both superior accuracy and high stability across random seeds.
The metadata curriculum ($1.822$\,m) performs worse than even the uniform baseline ($1.772$\,m), while the loss-based SPL curriculum ($2.003$\,m) exhibits severe instability (CV~=~$19.5\%$).

\begin{table}[t]
\centering
\caption{Multi-seed results (mean $\pm$ std, $n=3$ seeds). Best mean values in \textbf{bold}. $\downarrow$ = lower is better. CV = coefficient of variation.}
\label{tab:main_results}
\resizebox{\columnwidth}{!}{%
\begin{tabular}{lccccc}
\toprule
Method & planADE$\downarrow$ & planFDE$\downarrow$ & planAHE$\downarrow$ & predADE$\downarrow$ & CV \\
\midrule
Baseline & $1.772 \pm .134$ & $3.837 \pm .218$ & $.146 \pm .021$ & $1.700 \pm .205$ & 7.6\% \\
Meta cur. & $1.822 \pm .014$ & $3.996 \pm .216$ & $.142 \pm .010$ & $1.714 \pm .016$ & 0.7\% \\
\textbf{TracIn cur.} & $\mathbf{1.704 \pm .029}$ & $\mathbf{3.731 \pm .394}$ & $\mathbf{.133 \pm .019}$ & $1.731 \pm .095$ & \textbf{1.7\%} \\
Loss SPL & $2.003 \pm .391$ & $3.678 \pm .200$ & $.180 \pm .041$ & $\mathbf{1.633 \pm .069}$ & 19.5\% \\
Hybrid cur. & $1.766 \pm .069$ & $3.999 \pm .185$ & $.134 \pm .016$ & $1.707 \pm .047$ & 3.9\% \\
\bottomrule
\end{tabular}}
\end{table}

Table~\ref{tab:per_seed} provides per-seed breakdowns, showing that the TracIn curriculum outperforms the metadata curriculum in \textit{every} seed, with improvements of 0.145\,m, 0.122\,m, and 0.085\,m for seeds 3407, 42, and 2024 respectively.

\begin{table}[t]
\centering
\caption{Per-seed planning ADE (m, $\downarrow$). TracIn outperforms Meta in all 3 seeds. Best per-seed in \textbf{bold}.}
\label{tab:per_seed}
\resizebox{\columnwidth}{!}{%
\begin{tabular}{lccccc}
\toprule
Seed & Baseline & Meta cur. & TracIn cur. & Loss SPL & Hybrid cur. \\
\midrule
3407 & 1.917 & 1.832 & \textbf{1.687} & 1.728 & 1.772 \\
42   & \textbf{1.593} & 1.803 & 1.680 & 1.726 & 1.848 \\
2024 & 1.807 & 1.831 & \textbf{1.746} & 2.555 & 1.680 \\
\midrule
Mean & 1.772 & 1.822 & \textbf{1.704} & 2.003 & 1.766 \\
Std  & 0.134 & 0.014 & \textbf{0.029} & 0.391 & 0.069 \\
\bottomrule
\end{tabular}}
\end{table}

\subsection{Statistical Analysis}
\label{sec:statistics}

We conduct pairwise comparisons using paired $t$-tests across the three seeds (Table~\ref{tab:pairwise}).
The TracIn curriculum significantly outperforms the metadata curriculum ($p=0.021$, Cohen's $d_z=3.88$), a large effect size indicating the improvement is not only statistically significant but practically meaningful.

TracIn versus baseline is not statistically significant ($p=0.54$) due to the high variance of the baseline: seed 42 achieves $1.593$\,m while seed 3407 reaches $1.917$\,m, a $0.324$\,m swing.
However, the TracIn curriculum achieves both a lower mean and substantially lower variance (CV = $1.7\%$ vs.\ $7.6\%$) than the baseline, a practically important distinction for deployment reliability.
The metadata curriculum versus baseline comparison ($p=0.622$) confirms that metadata-based scoring provides no benefit.

\begin{table}[t]
\centering
\caption{Paired $t$-tests (planning ADE, $n=3$ seeds). Significant at $\alpha=0.05$ marked with~*.}
\label{tab:pairwise}
\begin{tabular}{lccc}
\toprule
Comparison & $\Delta$ ADE (m) & $p$-value & Cohen's $d_z$ \\
\midrule
TracIn vs.\ Meta  & $-0.117$ & $0.021^*$ & $3.88$ \\
TracIn vs.\ Base  & $-0.068$ & $0.540$ & $0.63$ \\
TracIn vs.\ SPL   & $-0.299$ & $0.296$ & $1.20$ \\
TracIn vs.\ Hyb   & $-0.062$ & $0.186$ & $1.71$ \\
Hybrid vs.\ Meta  & $-0.056$ & $0.291$ & $1.23$ \\
Meta vs.\ Base    & $+0.050$ & $0.622$ & $0.33$ \\
\bottomrule
\end{tabular}
\end{table}

\subsection{Training Dynamics}
\label{sec:training_dynamics}

\begin{figure}[t]
  \centering
  \includegraphics[width=\columnwidth]{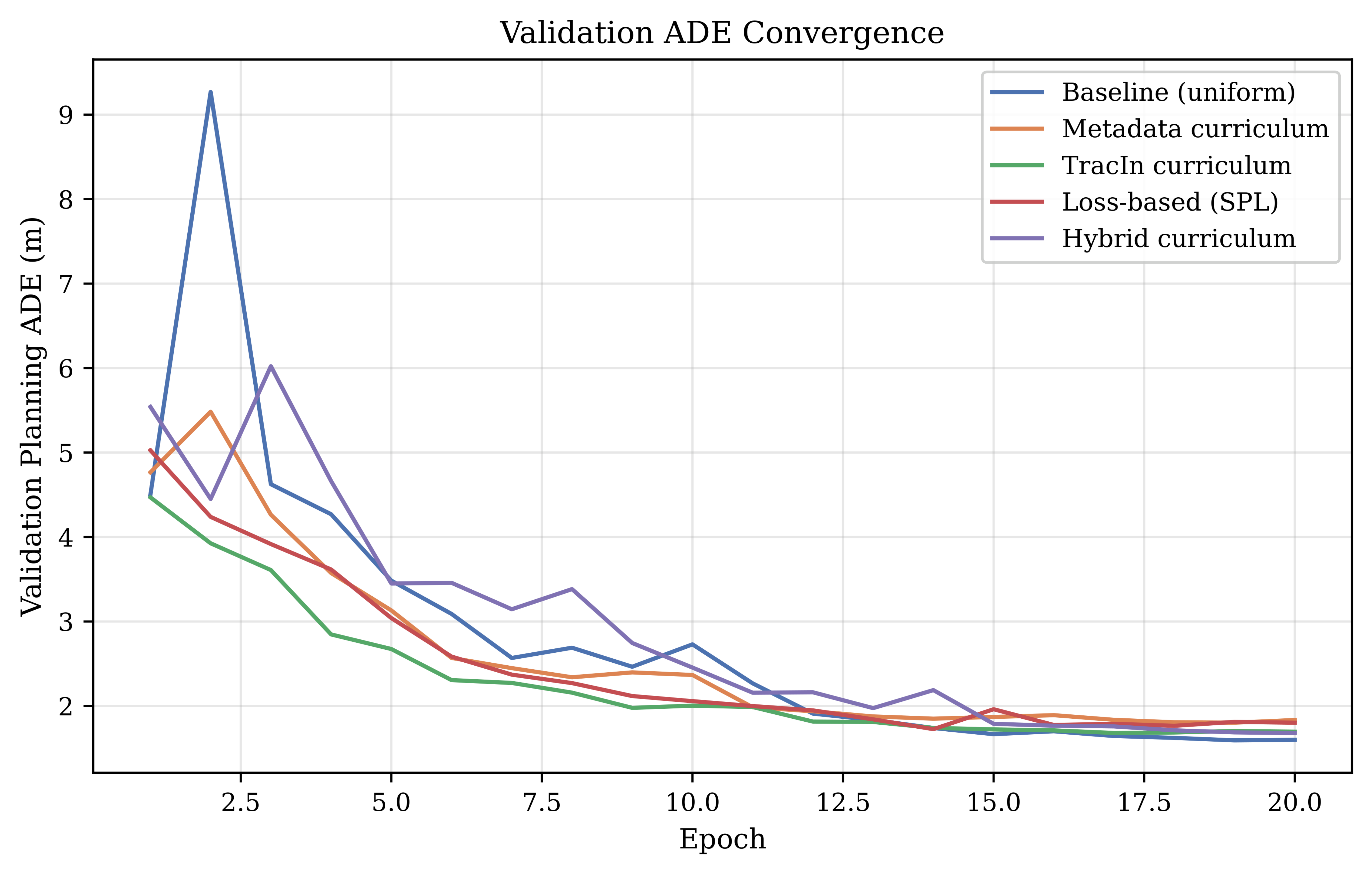}
  \caption{Validation planning ADE over training epochs for the best seed of each method. The TracIn curriculum (blue) achieves the lowest final ADE, converging below all other methods by epoch 15.}
  \label{fig:training_curves}
\end{figure}

Fig.~\ref{fig:training_curves} shows validation planning ADE as a function of training epoch for the best seed of each method.
The TracIn curriculum converges to a lower validation ADE than all other methods by epoch~15, and maintains this advantage through epoch~20.
The metadata curriculum and baseline exhibit similar convergence trajectories, consistent with their non-significant difference.
The loss-based SPL curve shows higher initial ADE due to its easy-first ordering, which delays exposure to informative scenarios.

\subsection{Failure Analysis}
\label{sec:failure}

\textbf{Hard selection degrades performance.}
Training on only the top 20\% of scenarios ranked by TracIn score produces a planning ADE of $3.687$\,m, more than $2\times$ worse than the full-data baseline.
High-TracIn samples are those most aligned with the current validation gradient, so restricting to these samples biases the training distribution toward a narrow region of scenario space.
Curriculum \textit{weighting} preserves full data coverage while adjusting emphasis, avoiding this distribution collapse.

\textbf{Loss-based SPL is unstable.}
The loss-based self-paced curriculum exhibits a coefficient of variation of 19.5\% across seeds.
Seed 2024 produces a catastrophic planning ADE of $2.555$\,m, while seeds 3407 and 42 achieve $1.728$\,m and $1.726$\,m respectively.
Training loss conflates intrinsic sample difficulty with model-specific uncertainty and optimization noise, making it an unreliable proxy for curriculum ordering.

\textbf{Hybrid scoring does not improve over TracIn.}
The hybrid rank-average of TracIn and metadata scores achieves $1.766\pm0.069$\,m, which does not significantly differ from TracIn alone ($p=0.186$).
Since metadata scores provide no useful signal for curriculum ordering (the metadata curriculum underperforms the baseline), combining them with TracIn via rank-averaging dilutes the effective gradient-based signal.
This suggests that when one scoring component is uninformative, combining it with an informative component via equal weighting is counterproductive.

\subsection{Multi-Metric Comparison}
\label{sec:multimetric}

\begin{figure}[t]
  \centering
  \includegraphics[width=\columnwidth]{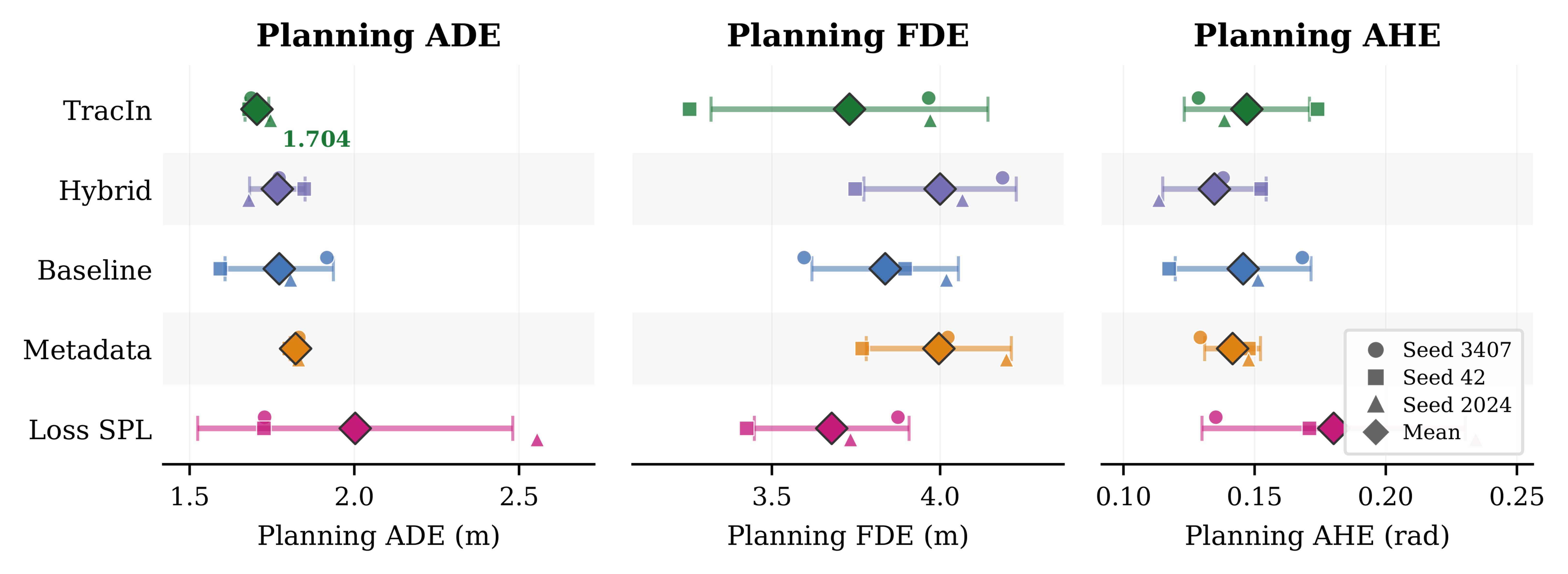}
  \caption{Dot-and-whisker comparison across five curriculum methods and three planning metrics. Diamonds: mean; shaped markers: individual seeds ($n{=}3$). Methods sorted by mean Planning ADE (lower is better). The TracIn curriculum achieves the best ADE with minimal seed-to-seed variation, while the loss-based SPL shows a large outlier.}
  \label{fig:multimetric}
\end{figure}

Fig.~\ref{fig:multimetric} shows per-seed performance across three planning metrics.
Methods are sorted by mean planning ADE (best at top); individual seed results (distinct markers) reveal the variance structure that summary statistics alone obscure.
The TracIn curriculum achieves the best mean ADE ($1.704$\,m) with tight seed clustering, while the loss-based SPL exhibits a catastrophic outlier at seed~2024 (ADE$=2.555$), confirming the instability reported in Table~\ref{tab:main_results}.

\section{Discussion}
\label{sec:discussion}

\textbf{Why gradients succeed where metadata fails.}
Metadata captures observable scenario properties that correlate with human-perceived difficulty but not with the model's learning dynamics.
A scenario with many nearby agents may be ``easy'' if interactions follow stereotypical patterns (Fig.~\ref{fig:scenario_demo}, bottom-right), while a geometrically simple scenario may be ``hard'' because it represents an underexplored region (Fig.~\ref{fig:scenario_demo}, top-left).
TracIn measures alignment between each sample's gradient and validation loss reduction, capturing model-specific difficulty that static features cannot encode ($\rho=-0.014$).

\textbf{Curriculum weighting vs.\ hard selection.}
Our results establish a critical practical insight: gradient-based scores are effective for importance weighting but counterproductive for subset selection.
Hard selection using TracIn's top 20\% removes diverse ``easy'' samples that provide necessary coverage of the scenario distribution, leading to overfitting on a narrow slice.
Curriculum weighting preserves this coverage while modulating emphasis---a strictly better approach when compute permits training on the full dataset.

\textbf{Connection to importance sampling.}
The curriculum weights in Eq.~\eqref{eq:curriculum} can be interpreted through the lens of importance-weighted empirical risk minimization~\cite{koh2017understanding}.
Standard ERM minimizes the training loss under the empirical distribution $p_{\text{train}}$, which may diverge from the effective deployment distribution.
TracIn-based weighting constructs an implicit reweighted distribution $p_w(z_i) \propto w_i \cdot p_{\text{train}}(z_i)$ that concentrates mass on samples whose gradients best reduce validation loss.
Under this view, the three-phase schedule corresponds to an annealing strategy: warm-up trains under the original $p_{\text{train}}$ to avoid premature bias, ramp-up gradually shifts toward $p_w$, and focus stabilizes at the target distribution.
This annealing is analogous to temperature schedules in simulated annealing---direct optimization under $p_w$ from the start risks overfitting to the initial (potentially noisy) gradient estimates, while gradual annealing allows the score quality to improve as the model trains.

\textbf{Practical considerations.}
TracIn scoring requires one forward-backward pass per sample: 46~min on a single GPU for 5{,}148 scenarios ($<$0.5\% of total compute).
For larger datasets, TracIn can be computed on a representative subset or parallelized across GPUs.

\textbf{Limitations.}
Our experiments use the nuPlan mini split (5{,}148 training scenarios).
While the statistical significance holds, the absolute performance difference is 0.117\,m ADE between TracIn and metadata curricula.
Validation on the full nuPlan dataset (130K+ scenarios) would strengthen the findings.
We evaluate only open-loop metrics; closed-loop simulation would provide a more complete assessment of planning quality.
The paired $t$-test with $n=3$ seeds has limited statistical power; increasing to $n \geq 5$ seeds would provide more robust inference.
Finally, our TracIn implementation uses a single checkpoint; multi-checkpoint scoring may yield richer temporal information about data utility.

\section{Conclusion}
\label{sec:conclusion}

We investigated data-centric methods for improving game-theoretic motion planning, comparing metadata-based, loss-based, and gradient-based curriculum strategies for GameFormer on nuPlan.
Our central finding is that TracIn gradient-similarity scoring produces curriculum orderings that significantly outperform metadata-based interaction-difficulty heuristics ($p=0.021$, Cohen's $d_z=3.88$), achieving the best mean planning ADE ($1.704$\,m) with low cross-seed variance (CV = 1.7\%).
The orthogonality between gradient-based and metadata-based scores ($\rho=-0.014$) reveals that gradient valuation captures model-specific training dynamics invisible to hand-crafted features.
We further identify a critical distinction between curriculum weighting (effective) and hard data selection (harmful), providing practical guidance for applying data valuation in planning systems.

Future work includes scaling to the full nuPlan dataset, extending to closed-loop evaluation, multi-checkpoint TracIn scoring, and applying gradient-based curriculum learning to other game-theoretic architectures.

\balance
\bibliographystyle{IEEEtran}
\bibliography{references}

@inproceedings{huang2023gameformer,
  title     = {{GameFormer}: Game-theoretic Modeling and Learning of Transformer-based Interactive Prediction and Planning for Autonomous Driving},
  author    = {Huang, Zhiyu and Liu, Haochen and Lv, Chen},
  booktitle = {Proceedings of the IEEE/CVF International Conference on Computer Vision (ICCV)},
  pages     = {3903--3913},
  year      = {2023}
}

@inproceedings{huang2023dtpp,
  title     = {{DTPP}: Differentiable Joint Conditional Prediction and Cost Evaluation for Tree Policy Planning in Autonomous Driving},
  author    = {Huang, Zhiyu and Karkus, Peter and Ivanovic, Boris and Chen, Yuxiao and Pavone, Marco and Lv, Chen},
  booktitle = {Proceedings of the IEEE International Conference on Robotics and Automation (ICRA)},
  pages     = {6806--6812},
  year      = {2024}
}

@inproceedings{dauner2023parting,
  title     = {Parting with Misconceptions about Learning-based Vehicle Motion Planning},
  author    = {Dauner, Daniel and Hallgarten, Marcel and Geiger, Andreas and Chitta, Kashyap},
  booktitle = {Conference on Robot Learning (CoRL)},
  pages     = {1268--1281},
  year      = {2023}
}

@inproceedings{cheng2024plantf,
  title     = {Rethinking Imitation-based Planners for Autonomous Driving},
  author    = {Cheng, Jie and Chen, Yingbing and Mei, Xiaodong and Yang, Bowen and Li, Bo and Liu, Ming},
  booktitle = {Proceedings of the IEEE International Conference on Robotics and Automation (ICRA)},
  pages     = {14123--14130},
  year      = {2024}
}

@inproceedings{pruthi2020tracin,
  title     = {Estimating Training Data Influence by Tracing Gradient Descent},
  author    = {Pruthi, Garima and Liu, Frederick and Kale, Satyen and Sundararajan, Mukund},
  booktitle = {Advances in Neural Information Processing Systems (NeurIPS)},
  volume    = {33},
  pages     = {19920--19930},
  year      = {2020}
}

@inproceedings{koh2017understanding,
  title     = {Understanding Black-box Predictions via Influence Functions},
  author    = {Koh, Pang Wei and Liang, Percy},
  booktitle = {Proceedings of the International Conference on Machine Learning (ICML)},
  pages     = {1885--1894},
  year      = {2017}
}

@inproceedings{ghorbani2019data,
  title     = {Data {Shapley}: Equitable Valuation of Data for Machine Learning},
  author    = {Ghorbani, Amirata and Zou, James},
  booktitle = {Proceedings of the International Conference on Machine Learning (ICML)},
  pages     = {2242--2251},
  year      = {2019}
}

@article{agarwal2017second,
  title     = {Second-Order Stochastic Optimization for Machine Learning in Linear Time},
  author    = {Agarwal, Naman and Bullins, Brian and Hazan, Elad},
  journal   = {Journal of Machine Learning Research},
  volume    = {18},
  number    = {116},
  pages     = {1--40},
  year      = {2017}
}

@inproceedings{feng2025tarot,
  title     = {{TAROT}: Targeted Data Selection via Optimal Transport},
  author    = {Feng, Lan and Nie, Fan and Liu, Yuejiang and Alahi, Alexandre},
  booktitle = {Proceedings of the International Conference on Machine Learning (ICML)},
  year      = {2025}
}

@article{lu2024activead,
  title     = {{ActiveAD}: Planning-Oriented Active Learning for End-to-End Autonomous Driving},
  author    = {Lu, Han and Jia, Xiaosong and Xie, Yichen and Liao, Wenlong and Yang, Xiaokang and Yan, Junchi},
  journal   = {arXiv preprint arXiv:2403.02877},
  year      = {2024}
}

@inproceedings{park2025galtraj,
  title     = {Generative Active Learning for Long-tail Trajectory Prediction via Controllable Diffusion Model},
  author    = {Park, Daehee and Surana, Monu and Desai, Pranav and Mehta, Ashish and John, Reuben MV and Yoon, Kuk-Jin},
  booktitle = {Proceedings of the IEEE/CVF International Conference on Computer Vision (ICCV)},
  year      = {2025}
}

@article{li2024datacentric,
  title     = {Data-Centric Evolution in Autonomous Driving: A Comprehensive Survey of Big Data System, Data Mining, and Closed-Loop Technologies},
  author    = {Li, Lincan and Shao, Wei and Dong, Wei and Tian, Yijun and Zhang, Qiming and Yang, Kaixiang and Zhang, Wenjie},
  journal   = {arXiv preprint arXiv:2401.12888},
  year      = {2024}
}

@article{caesar2021nuplan,
  title     = {{nuPlan}: A closed-loop {ML}-based planning benchmark for autonomous vehicles},
  author    = {Caesar, Holger and Kabzan, Juraj and Tan, Kok Seang and Fong, Whye Kit and Wolff, Eric and Lang, Alex and Fletcher, Luke and Beijbom, Oscar and Omari, Sammy},
  journal   = {arXiv preprint arXiv:2106.11810},
  year      = {2021}
}

@inproceedings{nayakanti2023wayformer,
  title     = {Wayformer: Motion Forecasting via Simple \& Efficient Attention Networks},
  author    = {Nayakanti, Nigamaa and Al-Rfou, Rami and Zhou, Aurick and Goel, Kratarth and Refaat, Khaled S. and Sapp, Benjamin},
  booktitle = {Proceedings of the IEEE International Conference on Robotics and Automation (ICRA)},
  pages     = {2980--2987},
  year      = {2023}
}

@inproceedings{bengio2009curriculum,
  title     = {Curriculum Learning},
  author    = {Bengio, Yoshua and Louradour, J{\'e}r{\^o}me and Collobert, Ronan and Weston, Jason},
  booktitle = {Proceedings of the International Conference on Machine Learning (ICML)},
  pages     = {41--48},
  year      = {2009}
}

@inproceedings{kumar2010selfpaced,
  title     = {Self-Paced Learning for Latent Variable Models},
  author    = {Kumar, M. Pawan and Packer, Benjamin and Koller, Daphne},
  booktitle = {Advances in Neural Information Processing Systems (NeurIPS)},
  volume    = {23},
  pages     = {1189--1197},
  year      = {2010}
}

@inproceedings{qiao2018curriculum,
  title     = {Automatically Generated Curriculum Based Reinforcement Learning for Autonomous Vehicles in Urban Environment},
  author    = {Qiao, Zhiqian and Muelling, Katharina and Dolan, John M. and Palanisamy, Praveen and Mudalige, Priyantha},
  booktitle = {Proceedings of the IEEE Intelligent Vehicles Symposium (IV)},
  pages     = {1233--1238},
  year      = {2018}
}

\end{document}